\documentclass[lettersize,journal]{IEEEtran}

\usepackage{amsmath,amsfonts}
\usepackage{algorithmic}
\usepackage{algorithm}
\usepackage{array}
\usepackage[caption=false,font=normalsize,labelfont=sf,textfont=sf]{subfig}
\usepackage{textcomp}
\usepackage{stfloats}
\usepackage{url}
\usepackage{verbatim}
\usepackage{graphicx}
\usepackage{cite}

\usepackage{graphicx}
\usepackage[normalem]{ulem}
\useunder{\uline}{\ul}{}

\usepackage{bm}

\begin{document}

\title{Physics-Driven AI Correction in Laser Absorption Sensing Quantification }

\author{Ruiyuan Kang$^{1}$,
        Panos Liatsis$^{2}$,~\IEEEmembership{Senior Member,~IEEE},
        Meixia Geng$^{1}$,
        and Qingjie Yang$^{1}$

\thanks{$^{1}$Ruiyuan.Kang, Meixia.Geng, and Qingjie.Yang@tii.ae are with Technology Innovation Institude, Abu Dhabi, UAE }
\thanks{$^{2}$Panos.Liatsis@ku.ac.ae is with Khalifa University, Abu Dhabi, UAE}
}



\maketitle

\begin{abstract}
Laser absorption spectroscopy (LAS) quantification is a popular tool used in measuring temperature and concentration of gases. It has low error tolerance, whereas current ML-based solutions cannot guarantee their measure reliability. In this work, we propose a new framework, SPEC, to address this issue. 
In addition to the conventional ML estimator-based estimation mode, SPEC also includes a Physics-driven Anomaly Detection module (PAD) to assess the error of the estimation. And a Correction mode is designed to correct the unreliable estimation. The correction mode is a network-based optimization algorithm, which uses the guidance of error to iteratively correct the estimation. A hybrid surrogate error model is proposed to estimate the error distribution, which contains an ensemble of networks to simulate reconstruction error, and true feasible error computation. A greedy ensemble search is proposed to find the optimal correction robustly and efficiently from the gradient guidance of surrogate model. The proposed SPEC is validated on the test scenarios which are outside the training distribution. The results show that SPEC can significantly improve the estimation quality, and the correction mode outperforms current network-based optimization algorithms. In addition, SPEC has the reconfigurability, which can be easily adapted to different quantification tasks via changing PAD without retraining the ML estimator.
\end{abstract}


\section{Introduction}
\IEEEPARstart{L}{aser} Absorption Sensing (LAS) is a widely used technique for gas concentration and temperature measurement \cite{wangLaserAbsorptionSensing2019}. It has been applied in many fields, such as combustion\cite{liuLaserAbsorptionSpectroscopy2019}, environmental monitoring \cite{dierksQuantificationMethaneGas2017}, etc. The principle of LAS is to shoot a laser beam through the target gas which can absorb the laser signal, and the absorption signal is collected by a detector, known as laser absorption spectrum $\mathbf{y} \in \mathcal{Y}$. According to Beer-lambert's law \cite{kangEmissionQuantificationPassive2022}, the absorption signal is related to gas temperature $T_{gas} $ and concentration $C_{gas}$, which is named as state $\mathbf{x}=\left\{T_{gas}, C_{gas}\right\} \in \mathcal{X}$. To estimate the state is the main task of LAS quantification.

Some physics-driven algorithms have been proposed to solve the spectroscopy quantification problem, such as two-color measurement \cite{hansonSpectroscopyOpticalDiagnostics2016a}, line-reversal measurement \cite{childsReviewTemperatureMeasurement2000}, etc. However, the application of these two methods need expert knowledge, and manual operation on spectral line picking and calibration \cite{goldensteinTwocolorAbsorptionSpectroscopy2013}, etc. All these prerequisites assure the accuracy of the quantification, but are time-consuming and inconvenient.

Recent advances in machine learning (ML) have provided much more efficient solution for the quantification problem. To realize this task, ML models are trained on pre-collected paired data, which are the absorption spectrum and the corresponding state. Through feeding the absorption spectrum into the ML model, and training the ML model under supervised learning style, ML models can thus estimate the gas concentration and temperature from the given spectrum \cite{ouyangQuantitativeAnalysisGas2019,zhouOnlineBlendtypeIdentification2019, hanIndustrialIoTIntelligent2020}.

However, the problem of these ML-based methods is that they cannot guarantee the reasonability of the quantification. This is because (1) ML models cannot even guarantee they thoroughly learn the training distribution, as they always have prediction errors, i.e., bias and variance \cite{pmlr-v151-wood22a}. (2) The training dataset cannot cover all the possible scenarios in deployment, because of measurement error, distribution shift, etc \cite{fang2020rethinking}. Nevertheless, as a serious scientific and engineering measuring tool, LAS has a low error tolerance for unreliable quantification.

To address this concern, some methods have been proposed for spectral application, such as using spectra from inhomogeneous temperature distribution \cite{kang2022intelligence} or with noise \cite{kang2022spatiallyresolved} to simulate the actual condition. Machine learning community also proposed some general solutions, such as training with physics-informed regularization \cite{karniadakisPhysicsinformedMachineLearning2021}, big model \cite{wengLargeLanguageModels2022}, special network architecture \cite{zhangRethinkingLipschitzNeural}, etc. These methods do relieve the issue but cannot solve it thoroughly, as they suppose the training data could represent and cover the real distribution, and ML models could learn the knowledge from the training data thoroughly, while both are ideal assumptions.

In this work, we propose a new framework, named as \textit{Surrogate-based Physical Error Correction (SPEC)}, to address the reliability concern in spectroscopy quantification. SPEC has two work modes: estimation mode and correction mode. For a given spectrum, the estimation mode is firstly activated, which in fact uses the existed ML estimator to estimate the state. But instead of making ML estimator to be as powerful as possible, we pursue to detect the unreliable state estimations via a Physics-driven Anomaly Detection module (PAD). PAD calculates an overall physical error $e \in \mathcal{E}$ according to the given estimation $\hat{\mathbf{x}}$ and spectrum $\mathbf{y}$. If the overall physical error $e$ is larger than a threshold $\epsilon$, the PAD will detect the estimation as anomaly, and the whole framework switches to correction mode. The correction mode will use the error information to correct the state, till the overall error is smaller than $\epsilon$. The correction mode is in fact solving the optimization problem:
\begin{equation}
    \min_{\hat{\mathbf{x}}} e(\hat{\mathbf{x}}; \mathbf{y}) \quad s.t. \quad \hat{\mathbf{x}} \in \mathcal{X}.
\end{equation}
The whole correction mode is based on a network-based optimization algorithm. A hybrid error surrogate model comprising the ensemble of networks and partial actual error computation, are trained with online data to estimate error distribution. A gradient-driven ensemble greedy search is proposed to exploit the surrogate model to search the corrected state. Meanwhile, Monte-Carlo sampling is used as the exploration strategy to collect data for updating the surrogate model. 

Our contributions are summarized as follows:
\begin{itemize}
    \item We harmoniously combine the ML-based estimator, physics-driven Anomaly Detection, and Network-based Optimization into a unified framework, to realize the reliable spectroscopy quantification. 
    \item The proposed physics-driven Anomaly Detection module can detect the anomaly reliably and also provide accurate error information for correction mode.The proposed Network-based Optimization algorithm can solve the correction problem efficiently and effectively, outperforms the existing network-based optimization algorithms.
    \item The proposed SPEC framework demonstrates its effectiveness on diverse deployment scenarios, and provide significant improvement on the reliability of the quantification than merely using ML-based estimator. Meanwhile, SPEC is reconfigurable that it can be easily adapted to different scenarios by changing PAD configuration without retraining.
\end{itemize}

\section{Related Work}
\label{sec:related_work}
\textbf{Machine Learning for Spectroscopy Quantification:}  Machine Learning has been a popular tool in spectroscopy quantification. Currently, the mean stream methodology in the field is to use supervised learning to learn the mapping between spectra and states, both classical machine learning algorithms, such as Support Vector Machine \cite{zhouOnlineBlendtypeIdentification2019, sun2021quantitative}, Random Forest \cite{khanehzar2022application}, MultiLayer Perceptron \cite{ouyangQuantitativeAnalysisGas2019, wang2021machine}, and Deep Learning algorithms, such as ConvNet \cite{boodaghidizaji2022characterizing, hanIndustrialIoTIntelligent2020}, LSTM+attention \cite{sun2022adaptively}, etc. have been applied. Without doubt, these models are fixed on the specific training set and cannot guarantee the reliability of the quantification in unseen test sets.
In order to enhance the reliability of the trained models, some research try to increase the spectral synthesis fidelity via adding instrument noise or simulating inhomogeneous temperature distribution \cite{kang2022spatiallyresolved, kang2022intelligence}. Some research utilizes transfer learning to fine tune the network on little actual experimental data or different wavebands \cite{tian2021retrieval, yi2021accurate}. These methods do enhance the reliability and expand the application scenarios of ML models, but they are still limited by the closed world of training and fine-tuning data, and cannot guarantee the reliability in all the possible scenarios in the open world.

\textbf{Anomaly Detection:} Anomaly detection \cite{yang2022generalized} is an important task in machine learning, which aims to detect the abnormal data from the normal data. Current Anomaly detection in Machine learning community is based on learning methodologies, including using GAN to distinguish the normal data and abnormal data \cite{xia2022gan}, finding the abnormal data via the disagreement between different models \cite{han2022outofdistribution}, using the reconstruction error of auto-encoder \cite{gong2019memorizing}, or learning (probabilistic) metrics which has the predefined threshold to distinguish anomaly \cite{gawlikowski2022advanced,hsu2020generalized}. Although these methods can detect whether a sample is Outside of Distribution (OoD) from the perspective of data distribution, but they cannot assess whether the estimation from  model is reliable (anomaly) or not. Furthermore, their detection cannot give reliable guidance on how to correct unreliable estimations. Therefore, in this work, we utilize the physical error to do the work of anomaly detection and provide authoritative guidance on how to correct the unreliable estimation.

\textbf{Neural Network for Optimization:} In fact, the correction mode we have is using neural network to solve the optimization problem, and the error from anomaly detection module is in equivalent to the objective function of the optimization problem. The prerequisite for doing so is that optimization objective function, i.e., physical error function, should be differentiable. In OptNet or Iterative Neural Network \cite{amosOptNetDifferentiableOptimization,chun2023momentumnet}, it requires the physical process is naturally differentiable and can be embedded as a block of neural network. However, in most engineering applications including spectroscopy quantification, the physical process is indifferentiable, as the physical models could be a hybrid of differentiable equations and non-differentiable database and maps. 

Therefore, differentiable surrogate model, i.e., neural network, is needed to approximate the objective function, and then utilized to optimize the state. Such a paradigm is known as predict and optimize \cite{mandiSmartPredictandOptimizeHard2019}. There are different ways to utilize the surrogate model, including refocusing the existing ML model with the guidance of surrogate model \cite{fannjiang2020autofocused, kangSelfValidatedPhysicsEmbeddingNetwork2022}, training a generative model to generate new states (Tandem Network) \cite{liuTrainingDeepNeural2018, guan2023machine}, adding one neural layer with input of dummy vector (latent vector) to the surrogate model to optimize the input of surrogate model \cite{peurifoyNanophotonicParticleSimulation2018a,chenNeuralOptimizationMachine2022}, or treating the surrogate model as an energy-based model\cite{lecunEnergyBasedModelsDocument2007} to directly search for the global minimum via different search strategies \cite{ villarrubiaArtificialNeuralNetworks2018,vaezinejadHybridArtificialNeural2019}. In addition, a close field for such optimization problems is reinforcement learning\cite{haarnojaSoftActorCriticAlgorithms2018,moerland2023model}, where the surrogate model is equivalent to the world/critic model. However, the emphasis of reinforcement learning is to training a policy model to tackle all potential conditions, while the optimization problem faced here is to find reliable state for a given spectrum.

It is notable that considering surrogate model cannot be perfect, in this work, we insist on assessing the corrected state by Physics-driven Anomaly Detection Module. This is different from the general surrogate-model-based optimization \cite{li2019surrogate}.
  
\section{Proposed SPEC Framework}
\label{SPEC}
We firstly explain the notation convention:
  Ordinary letters, such as $x$ or $X$, represent  scalars or functions with scalar output.
  Bold letters, such as $\mathbf{x}$ or $\mathbf{X}$, represent vectors or functions with vector output. 
  The $i$-th element of $\mathbf{x}$  is denoted by $\mathbf{x}[i]$, while the first $k$ elements of $\mathbf{x}$ by $\mathbf{x}[1:k]$. Different $\mathbf{x}$ is indicated by $\mathbf{x}_i$. The data updated in each iteration $t$ is denoted by $\mathbf{x}^{(t)}$.
  We use $|\mathbf{x}|$,  $\|\mathbf{x}\|_1$ and $\|\mathbf{x}\|_2$ to denote the dimension, $l_1$-norm and $l_2$-norm of the vector $\mathbf{x}$.

\begin{figure*} [t]
    \centering
    \includegraphics[width=0.9\textwidth]{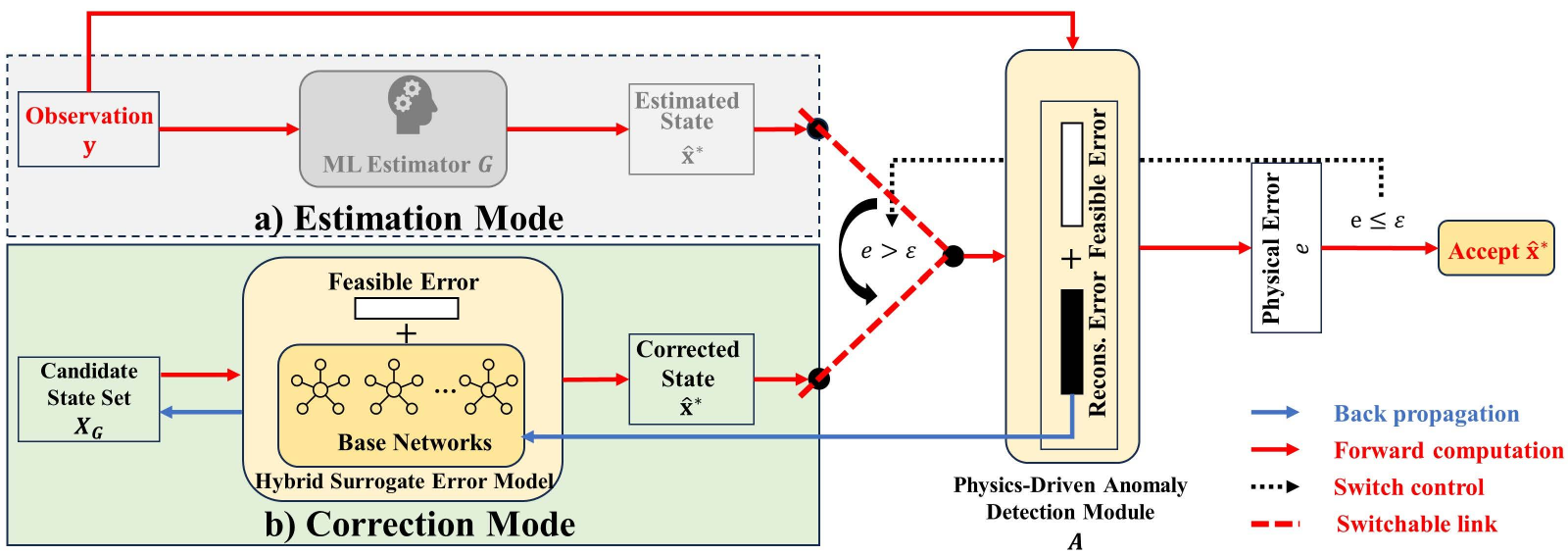}
    \caption{The Workflow of SPEC: ML estimator $G$ gives first estimation, which is fed to physics-driven Anomaly Detection (PAD) Module $A$ to calculate actual error $e$. If $e > \epsilon$, correction mode is activated. The error estimated by hybrid surrogate error model is used to guide the optimization of candidate dataset $X_\textmd{C}$. The state $\hat{\mathbf{x}}^* \in X_\textmd{C}$ leading to minimal estimated error is feed to PAD for evaluation. The process is terminated till $e(\hat{\mathbf{x}}^*) \leq  \epsilon$ or the iteration budget $T$ is exhausted.}
    \label{fig:workflow}
  
  \end{figure*}

 Figure\ref{fig:workflow} demonstrates the general workflow of SPEC. SPEC has two work modes: estimation mode and correction mode, its operational procedures are as follows: (1) Estimation mode is first activated, the trained ML estimator $G$ is used to give the first estimation of the state $\hat{\mathbf{x}}^{(0)}$. (2) The first guess is assessed by the Physics-driven Anomaly Detection module (PAD) $A$. $A$ produces the overall physical error $e$ as the assessment metric. If the error is no more than the predefined threshold $\epsilon$, the estimation is regarded as successful and the process is terminated. (3) Otherwise, the first estimation is detected as an anomaly and the correction mode is activated.  Correction Mode is a gradient-based Optimization, where an ensemble network-based model, hybrid error estimator $H$, is used to learn and simulate the error $\hat{e}$. And a set of randomly initialized states $X_\textmd{C} \subset \mathcal{X}$ are optimized via the back propagation of $H$ to reduce the estimated error. (4) The corrected state $\hat{\mathbf{x}} \in X$ leading to the minimal estimated error $\hat{e}$ is fed to PAD to calculate actual overall error $e(\hat{\mathbf{x}})$. The process is iterated till$ e(\hat{\mathbf{x}})$  no more than the predefined threshold $\epsilon$ or the predefined iteration number is triggered. The detailed description of each component is given in the following subsections.

\subsection{Estimation Mode}
\label{sec:EM}
The estimation mode has no difference with conventional applications of ML model: A existing ML estimator $G$ is  deployed to give the first estimation of the state $\hat{\mathbf{x}}^{(0)}$ according to the given spectrum $\mathbf{y}$, i.e., $G: \mathcal{Y} \rightarrow \mathcal{X}$. In principle, the model can be any regressive machine learning model. The model is shaped offline with the training dataset $D_{offline} = \{(\mathbf{y}_i, \mathbf{x}_i)\}_{i=1}^K$, where $\mathbf{y}_i$ is the $i$-th measured spectrum and $\mathbf{x}_i$ is the corresponding state (temperature and concentration of a given species).

\subsection{Physics-driven Anomaly Detection}
\label{sec:PAD}
The Physics-driven Anomaly Detection (PAD) module $A$ is used to assess the estimation of the state $\hat{\mathbf{x}}$. The assessment contain two parts: reconstruction error and feasible error. 

\textbf{Reconstruction Error} $e_{\textmd{R}}$ is based on the physical forward model $F$, which can transform the state $\mathbf{x}$ to the spectrum $\mathbf{y}$, i.e., $F: \mathcal{X} \rightarrow \mathcal{Y}$. Therefore, for a given estimated state $\hat{\mathbf{x}}$, one can calculate the corresponding estimated spectrum $\hat{\mathbf{y}}$, and then compare it with the measured spectrum $\mathbf{y}$ to get the reconstruction error $e_{\textmd{R}}$:
\begin{equation}
\label{eq:reconstruction_error}
    e_{\textmd{R}}(\hat{\mathbf{x}},\mathbf{y}) = \| F(\hat{\mathbf{x}}) - \mathbf{y} \|_2 .
\end{equation}

However, merely the reconstruction error $e_{\textmd{R}}$ is not enough to assess the quality of state estimation. The reason is spectrum quantification is an ill-posed problem, where the solution is not unique. To address this issue, we introduce the feasible error.

\textbf{Feasible Error} $\mathbf{e}_{\textmd{F}}$ is used to check whether the estimation of state $\hat{\mathbf{x}}$ is in the predefined feasible domain. It is understoodable that  the temperature and concentration of species are thoroughly different in different applications, such as in the flame or the atmosphere. Therefore, one can utilize this \textit{prior} to set different feasible domains for different applications. In our case, the feasible domain is defined as:
\begin{equation}
    \mathcal{X}_{\textmd{F}} = \left\{ \mathbf{x} \in \mathcal{X} \mid \mathbf{x}_{\textmd{min}} \leq \mathbf{x} \leq \mathbf{x}_{\textmd{max}} \right\} ,
\end{equation}
where $\mathbf{x}_{\textmd{min}}$ and $\mathbf{x}_{\textmd{max}}$ are the lower and upper bounds of the feasible domain, respectively. The feasible error  $\mathbf{e}_{\textmd{F}}[i]$ for the state element $\mathbf{x}[i]$ is defined as:
\begin{equation}
  \label{boundary error}
  \begin{split}
    \mathbf{e}_{\textmd{F}}(\hat{\mathbf{x}})[i]  = & \max\left( \frac{\hat{\mathbf{x}}[i]-x_{\textmd{min}}[i]}{x_{\textmd{max}}[i]-x_{\textmd{min}}[i]}-1, 0\right)+ \\
    &\max\left(- \frac{\hat{\mathbf{x}}[i]-x_{\textmd{min}}[i]}{x_{\textmd{max}}[i]-x_{\textmd{min}}[i]},0 \right) ,
  \end{split}
\end{equation}
where $\mathbf{e}_{\textmd{F}}[i]$ is the feasible error of the $i$-th element of the state $\hat{\mathbf{x}}$, and $\hat{\mathbf{x}}[i]$ is the $i$-th element of the estimated state $\hat{\mathbf{x}}$. This equation elaborates that once the estimated state element $\hat{\mathbf{x}}[i]$ exceeds the feasible domain, the corresponding estimation error $\mathbf{e}_{\textmd{F}}[i]$ will be greater than zero. In the application herein, $e_{\textmd{F}}$ has two elements, which are corresponding to the assessment of temperature and concentration, respectively. 
By introducing feasible error, we can alleviate the ill-posed problem, narrow the correction domain, and navigate the correction direction.

Finally, the overall physical error $e$ is defined as the weighted sum of both reconstruction error $e_{\textmd{R}}$ and feasible error $\mathbf{e}_{\textmd{F}}$:
\begin{equation}
  \label{overall error}
    e (\hat{\mathbf{x}}, \mathbf{y}) = w_{\textmd{R}} e_{\textmd{R}} +w_{\textmd{F,1}} \mathbf{e}_{\textmd{F}}[1]+w_{\textmd{F,2}} \mathbf{e}_{\textmd{F}}[2] .
\end{equation}
In our case, the weights are set to be 1, because all there components considered in the overall error are all important and have similar error slot. 

For a given spectrum $\mathbf{y}$, the dependence of $e$ on this spectrum is constant. Therefore, we can simply the expression of overall physical error $e$ as a function only depends on the estimated state $\hat{\mathbf{x}}$:
\begin{equation}
  e(\hat{\mathbf{x}}, \mathbf{y}) =e(\hat{\mathbf{x}}|\mathbf{y}) =  e(\hat{\mathbf{x}}).
\end{equation}
If the overall physical error $e$ is no more than the predefined threshold $\epsilon$, the estimation is regarded as successful and the process is terminated. Otherwise, the estimation is detected as an anomaly and the correction mode is activated.

\subsection{Correction Mode}
\label{sec:CM}
The correction mode is a network-based optimization algorithm. It has four steps which make an iteration cycle:(1) Estimation: Training surrogate model with online collected data; (2) Exploitation: Search the corrected state $\hat{\mathbf{x}}^*$ via greedy ensemble search according to the guidance of surrogate model; (3)Assessment: Evaluate the corrected state $\hat{\mathbf{x}}^*$ by PAD; (4) Exploration: Collect more data to update the surrogate model.

\subsubsection{Hybrid Surrogate Error Model} Considering PAD $A$ contains indifferentiable physical forward model, we cannot directly calculate the gradient from it. A common operation is to simulate the behavior of PAD directly via a differentiable surrogate model \cite{han2022surrogate}. However, it is notable that PAD also contains the differentiable feasible error $e_{\textmd{F}}$, so wrapping all error information into a black-box surrogate model does not only increase the estimation uncertainty but also lose the guidance from error structure \cite{hinton2022forward}.

To address this issue, we propose a hybrid surrogate error model to respectively process these two error components. Utilizing the advantage of ensemble learning\cite{hanEnsembleDeepLearning2022}, we use an ensemble of $L$ base neural networks to provide robust and accurate estimation of the distribution of reconstruction error $e_{\textmd{R}}$. In addition, we directly calculate the feasible error $\mathbf{e}_{\textmd{F}}$ from the estimated state $\hat{\mathbf{x}}$ without the simulation of surrogate models.

Each base neural network herein is fully connected with a mapping function $\bm\phi(\mathbf{x}, \mathbf{w}):  \mathcal{R}^{|\mathbf{x}|} \times  \mathcal{R}^{|\mathbf{w}|} \rightarrow   \mathcal{R}$.
The network weights are stored in the vector $\mathbf{w}$. 
We train $L$ individual  base networks sharing the same  architecture, while obtain the final prediction using an average combiner. 
As a result, given a state estimation $\hat{\mathbf{x}}$, the estimate of reconstruction error  is computed by
\begin{equation}
\hat{e}_{\textmd{R}}\left(\hat{\mathbf{x}}, \{\mathbf{w}_i\}_{i=1}^L\right)= \frac{1}{L}\sum_{i=1}^L \bm\phi\left(\hat{\mathbf{x}}, \mathbf{w}_i\right),
\end{equation}
and thus, the overall physical error is approximated by
\begin{equation}
\label{error_estimation}
\begin{split}
\hat{e}\left(\hat{\mathbf{x}}, \{\mathbf{w}_i\}_{i=1}^L\right)= & \underbrace{ \frac{1}{L}\sum_{i=1}^L  \phi \left(\hat{\mathbf{x}}, \mathbf{w}_i\right)}_{\textmd{approximated reconstruction error}}+  \\ & \underbrace{\sum_{j=1}^2 e_{\textmd{F}}(\hat{\mathbf{x}})[j]}_{\textmd{true feasible error}}.
\end{split}
\end{equation}
We refer to Eq. (\ref{error_estimation}) as a hybrid surrogate error model including both approximated and true error evaluation.

The   weights of the base neural networks $\{\mathbf{w}_i\}_{i=1}^L$  are trained using a set of online collected state-error pairs, e.g., $D=\{\left(\hat{\mathbf{x}}_i, e_{\textmd{R},i}\right)\}_{i=1}^Z$. In our implementation,  bootstrapping sampling   \cite{mooney1993bootstrapping} is adopted to train each base neural network independently, by minimizing a distance loss between the estimated and collected implicit errors, as
\begin{equation}
\label{train_error}
\min_{ \mathbf{w}_i } \mathbb{E}_{(\hat{\mathbf{x}}, e_{\textmd{R}})\sim D } \left[\textmd{dist}\left( \bm\phi\left(\hat{\mathbf{x}}, \mathbf{w}_i\right)  , e_{\textmd{R}}\right)\right].
\end{equation}
Where distance function $\textmd{dist}(\cdot, \cdot)$ is defined as Euclidean Distance in our implementation.

\subsubsection{Greedy Ensemble Search} Massive ways of utilizing the surrogate error model to search the corrected state $\hat{\mathbf{x}}^*$ are available, as we discussed in Section \ref{sec:related_work}. However, inheriting the existed ML estimator $G$ \cite{fannjiangAutofocusedOraclesModelbased,kangSelfValidatedPhysicsEmbeddingNetwork2022} or training a new one \cite{peurifoyNanophotonicParticleSimulation2018a,guan2023machine} is computationally expensive;  additionally, starting from one fixed state to search the corrected state $\hat{\mathbf{x}}^*$ is inefficient and may be trapped in local minimum. As for directly searching the state \cite{vaezinejadHybridArtificialNeural2019} leading to lowest estimated error $\hat{e}$,  it could lead to overfitting because of the inconsistency of true error and estimated error \cite{pmlr-v151-wood22a}.

In this study, we propose a greedy ensemble search, which searches the state leading to lowest estimated error $\hat{e}$ among a set of candidate states $X_\textmd{C} \subset \mathcal{X}$, and utilizes the gradient information of the hybrid surrogate error model $H$ to guide the search.

The initial candidate set $X$ is generated by Monte-Carlo Search , i.e., ${X_\textmd{C}}^{(0)}=\{\hat{\mathbf{x}}_i\}_{i=1}^{N_\textmd{C}} \subset \mathcal{X}_\textmd{F}$, where $N_\textmd{C}$ is the number of candidate states.

They can be updated iteratively by
\begin{equation}
\label{update_candidate}
{X_\textmd{C}}^{(t)}= \arg\min_{X_\textmd{C} \subset \mathcal{X}} \mathbb{E}_{\mathbf{x}_\textmd{C} \in {X_\textmd{C}}}\left[\hat{e}\left( \mathbf{x}_\textmd{C},  \left\{\mathbf{w}_i^{(t-1)}\right\}_{i=1}^L\right) \right],
\end{equation}
where the base networks from  the last  iteration are used, and we add the subscript  $t-1$ to the weights of the error network  for emphasizing. 
Finally,  among the candidates in $X_\textmd{C}^{(t)}$, we select the following  state
\begin{equation}
\label{candidate1}
\hat{\mathbf{x}}_{\textmd{C}}^{(t)} = \arg\min_{ \hat{\mathbf{x}}_\textmd{C} \in {X_\textmd{C}}^{(t)} }  \hat{e} \left(\hat{\mathbf{x}}_\textmd{C},  \left\{\mathbf{w}_i^{(t-1)}\right\}_{i=1}^L\right),
\end{equation}
to be the candidate state, and fed to PAD for assessment (Eq. (\ref{overall error})), resulting in the state-error pair $\left(\hat{\mathbf{x}}_{\textmd{C}}^{(t)}, \mathbf{e}_{\textmd{C}}^{(t)}\right)$. 

However, such an updating style (Eq.\ref{update_candidate}) may lead to neural collapse \cite{zhuTiCoTransformationInvariance2022}, i.e., all states in $X_\textmd{C}$ are converged to one state. To avoid this, we add an error element, diversity error $e_{\textmd{D}}$, into the update rule (Eq.\ref{update_candidate}). The diversity error is defined as:
\begin{equation}
\label{diffusion_error}
e_{\textmd{D}}(X_{\textmd{C}}) =\frac{\max(0.288c_1-\sigma(\frac{X_\textmd{C}-\mathbf{x}_{min}}{\mathbf{x}_{max}-\mathbf{x}_{min}}), 0)}{0.288c_1}*\frac{\epsilon}{c_2},
\end{equation}
where, $\sigma(\cdot)$ is the standard deviation of the normalized candidate set  bounded by feasible domain. 0.288 is the standard deviation of a uniform distribution bounded by $[0, 1]$, which is the same range of normalized candidate set. 
The diversity error elaborates the encouragement of the diversity of the candidate set $X_\textmd{C}$ (represented by its standard deviation). When the standard deviation of the normalized candidate set is small, diversity error will increase. $c_1$ is used to decide when to activate the diversity error, $c_2$ controls the magnitude of the diversity error. They are respectively 5 and 2 in our implementation. Accordingly, we modified Eq. (\ref{update_candidate}) as:
 \begin{equation}
  \begin{split}
    \label{update_candidate1}
    {X_\textmd{C}}^{(t)}= & \arg\min_{X_\textmd{C} \subset \mathcal{X}} \\& \underbrace{\mathbb{E}_{\mathbf{x}_\textmd{C} \in {X_\textmd{C}}}    \left[\hat{e}\left( \mathbf{x}_\textmd{C},  \left\{\mathbf{w}_i^{(t-1)}\right\}_{i=1}^L\right) \right]}_{\textmd{Estimated Overall Error}}  + \underbrace{e_{\textmd{D}}(X_{\textmd{C}})}_{\textmd{diversity error}}.
  \end{split}
 \end{equation}

Such a greedy ensemble search can significantly improve the search efficiency and avoid being trapped in local minimum, because of the search is started from multiple states and diversity error encourages the diversity of the candidate set. Meanwhile, the gradient and error structure information of the hybrid surrogate error model is used to guide the search, which can further improve the search efficiency. In addition, because we can control the update times and the number of candidates in each iteration, the overfitting problem can be alleviated.

\subsubsection{Assessment and Data Collection} \label{sec:assessment}
At the very beginning of the correction mode,  in order to activate the training of hybrid surrogate error model $H$, we use Monte-Carlo sampling to sample one batch size, $N$ states from the feasible domain, and assess them via PAD, to collect state-reconstruction error pairs, i.e.,
\begin{equation}
  \begin{split}
    & \hat{\mathbf{x}}_{\textmd{M}} =\mathcal{U}[\mathbf{x}_{min}, \mathbf{x}_{max}]
    \\ & e_{\textmd{R,M}} = e_{\textmd{R}}\left(\hat{\mathbf{x}}_{\textmd{M}}\right), \\ & e_{\textmd{F,M}} = e_{\textmd{F}}\left(\hat{\mathbf{x}}_{\textmd{M}}\right), \\ & e_{\textmd{M}} = e_{\textmd{R,M}}+e_{\textmd{F,M}}.
  \end{split}
\end{equation}

The state-reconstruction error pairs are stored into the data buffer $D^{(t)}$  for training the surrogate error model, $t$ is the iteration number, and $t=0$ herein, i.e., 
\begin{equation}
  \label{initial data buffer}
  D^{(0)} = \left\{ \left(\hat{\mathbf{x}}_{\textmd{M},i}, e_{\textmd{R,M},i}\right) \right\}_{i=1}^N,
\end{equation}

Next, in each iteration $t$, the corrected state ${\hat{\mathbf{x}}_{\textmd{C}}}^{(t)}$  generated from greedy gradient-based search are also assessed by PAD module. If the overall physical error is less than the feasibility threshold $\epsilon$, the correction mode is stopped and the corrected state is regarded as the feasible state $\hat{\mathbf{x}}^*$. Otherwise, Monte-Carlo sampling will generate one state ${\hat{\mathbf{x}}_{\textmd{M}}}^{(t)}$ to query PAD again. The corresponding two new state-error pairs are stored into the data buffer $D^{(t)} = D^{(t-1)}  \cup \left(\hat{\mathbf{x}}_{\textmd{C}}^{(t)}, e_{\textmd{R,C}}^{(t)}\right)  \cup \left(\hat{\mathbf{x}}_{\textmd{M}}^{(t)}, e_{\textmd{R,M}}^{(t)}\right)$, where  $e_{\textmd{R,C}}$ is the reconstruction error of the corrected state ${\hat{\mathbf{x}}_{\textmd{C}}}^{(t)}$.
Then, the base neural network weights  $ \mathbf{w}_i^{(t-1)} $   obtained from the previous iteration are  further fine-tuned  using the two added samples $ \left(\hat{\mathbf{x}}_{\textmd{C}}^{(t)}, e_{\textmd{R,C}}^{(t)}\right)$ and $\left(\hat{\mathbf{x}}_{\textmd{M}}^{(t)}, e_{\textmd{R,M}}^{(t)}\right)$, as well as $N$ examples sampled from the previous training set $D^{(t-1)}$.

\subsubsection{Balance between training and exploitation}
When training the base neural networks for reconstruction error estimation, in addition to setting maximum epochs $T_e$ in every iteration, early stopping of the training is enforced when the training loss in Eq. (\ref{train_error}) is smaller than a predefined threshold $\epsilon_{e}$. 
As a result, a higher number $n_e$ of early stopped base neural networks indicates a potentially more accurate error estimation.  
This strengthens the confidence in exploitation via greedy ensemble search in the next iteration.
In other words, when the base neural network are not sufficiently well trained, it is not recommended putting much effort in updating the candidate state by greedy Ensemble search, as it may be misled by the inaccurate error estimation.
Therefore, we set the maximum iteration number $T_G$ for updating greedy ensemble search in proportional to $n_e$, i.e., $T_G = \delta_G\lfloor \frac{2n_e}{L}+1\rfloor$, where $\delta_G$ is training frequency coefficient, which is a hyperparameter to be tuned.

\subsection{Implementation Details}
\textbf{Estimation Mode} Because of our emphasis on correction mode, we merely use an ordinary VGG-13 \cite{simonyanVeryDeepConvolutional2015} to be the backbone of ML estimator $G$. In order to process 1D spectral signal, we change all 2D convolution layers into 1D ones, and remove batch normalization operations for regression task. In addition, we add adaptive pooling layer to the end of convolution layers, so that the ML estimator can process spectral signals with different lengths. The acquired feature vectors acquired from adaptive pooling layer are fed to two dense layers, with the output dimension of 256 and 2, respectively. The activation function of the first dense layer is ReLU, and the second one is linear. The output of the second dense layer is the estimated state $\hat{\mathbf{x}}$.
  
\begin{figure}[hbtp!]
  \centering
  \includegraphics[width=1.\columnwidth]{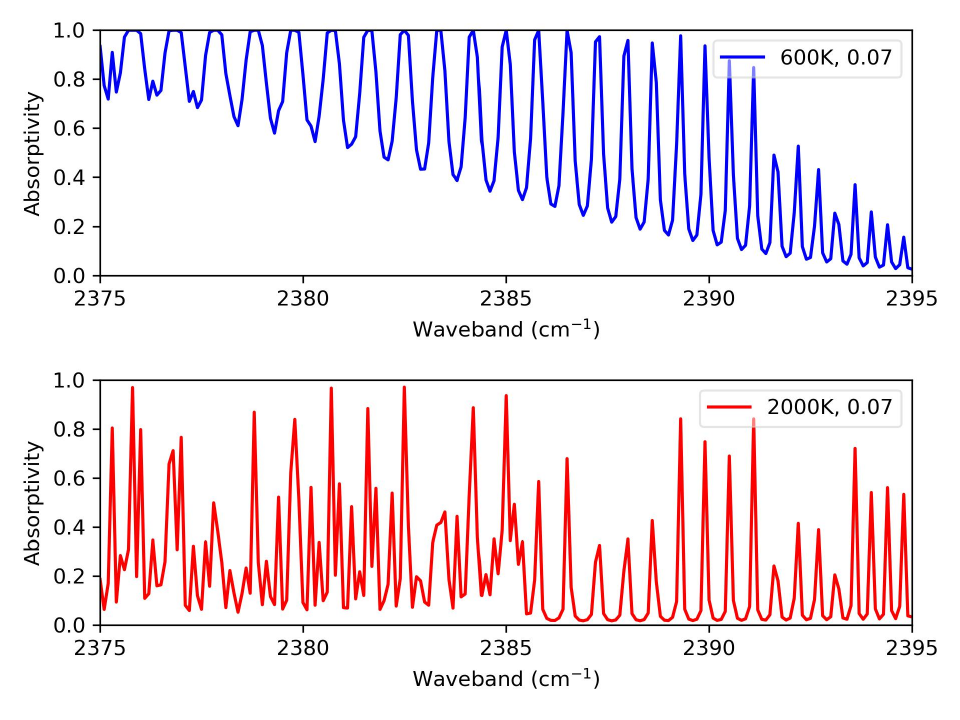}
  \caption{ CO$_2$ absorption spectra at 600 K (top) and 2000 K (bottom), with a mole fraction of 0.07.}
  \label{fig: waveband sensitivity}
\end{figure}

In order to train ML estimator $G$, we use HITEMP database \cite{rothmanHITEMPHightemperatureMolecular2010} and a well-recognized physical forward model: Radis\cite{pannierRADISNonequilibriumLinebyline2019}, to generate the dataset $D_{offline}$. The molecule chosen for this investigation is carbon dioxide (CO$_2$). The states, i.e., temperature and mole fraction, were assigned randomly in the range of 600-2000 K, and 0.05-0.07, respectively. The waveband selected is 2375-2395 $cm^{-1}$, since it is sensitive to temperature changes of CO$_2$ (Fig.\ref{fig: waveband sensitivity}). By setting an interval of 0.1 cm$^{-1}$, the generated spectrum has 200 dimensions, i.e.,$|\mathbf{y}|=200$. We synthesize 10,000 samples in total, and split it into training, validation and test sets with the partitions of 70\%, 15\%, 15\%, respectively.

\textbf{Physics-Driven Anomaly Detection Module} We use Radis as the selected physical forward model $F$ to calculate the reconstruction error (Eq. \ref{eq:reconstruction_error}). Of course, one can also choose any other spectral forward simulation platform, such as HAPI \cite{kochanovHITRANApplicationProgramming2016}. The feasible domain used for calculating feasible error $\mathbf{e}_\textmd{F}$ is changed according to the test scenarios, which will be described in the following experiments.
  
\textbf{Correction Mode}. In correction Mode, we use a simple MultiLayer Perceptron to be the base network $\phi$, which has the architecture of $2 \rightarrow 512 \rightarrow 1024 \rightarrow 512 \rightarrow 1$. The activation function of the first three layers is ReLU, and the last one is Sigmoid for producing non-negative error value, and a coefficient of 2 is timed to the output  in order to amplify the error estimation slot. In total, four base networks are trained via bootstrapping sampling. In one iteration, we train base networks with maximal epochs $T_e=40$, and the batch size $N$ of training base networks is 32, accordingly, we first do 32 Monte-Carlo sampling to collect $D^{(0)}$ so that activating the training of base network. As for greedy ensemble search, we set the number of candidate states ${N_\textmd{C}}=128$. The training frequency coefficient $\delta_G$ is set to be 1. The learning rate for base network and greedy ensemble search are respectively 1e-4, and 2.5e-2, $\epsilon_{e}$ for early stopping is set as 1e-4

The pseudocode of the proposed SPEC is shown in Algorithm \ref{alg1}, which displays the whole process of SPEC. 
\begin{algorithm}[t]
  \caption{SPEC}
  \label{alg1}
  \begin{algorithmic}[1]
  \REQUIRE{A spectrum $\mathbf{y}$, existed ML estimator $G$, physical anomaly detection Module $A$, base networks $\left\{\theta_i\right\}_{i=1}^L$, feasibility threshold $\epsilon>0$, training frequency coefficient $\delta_G$,  maximal iteration numbers $T$ and maximal epoch $T_e$, early stopping threshold $\epsilon_e>0$, batch size $N$ }  
  \ENSURE{An acceptable state $\hat{\mathbf{x}}^*$ with $e(\hat{\mathbf{x}}^*) \leq \epsilon$}
  \STATE \textbf{ESTIMATION MODE}
  \STATE  $\hat{\mathbf{x}}_{\textmd{0}} = G(\mathbf{y}), \quad e_{\textmd{R}}= e_{\textmd{R}}(\hat{\mathbf{x}}), \quad e_{\textmd{F}}= e_{\textmd{F}}(\hat{\mathbf{x}})$
  \STATE $\hat{\mathbf{x}}^*= \hat{\mathbf{x}}_{\textmd{0}}$
  \IF { $e_{\textmd{R}} +e_{\textmd{F}}\leq \epsilon$ }
  \STATE Stop the algorithm 
  \ENDIF
  \STATE \textbf{CORRECTION MODE}
  \STATE \textbf{Initialize:} iteration index  $t=0$, initial base neural network weights $ \left\{\mathbf{w}_i^{(0)}\right\}_{i=1}^L$,   number of early stopped base neural networks $n_e=0$, initial data buffer $D^{(0)}=\left\{ \left(\hat{\mathbf{x}}_{\textmd{M},i}, e_{\textmd{R,M},i}\right) \right\}_{i=1}^N$
  
  \FOR{$t \leq T$}
    \STATE Update $ \left\{\mathbf{w}_i^{(t)}\right\}_{i=1}^L$ by training each base neural network using  $ D^{(t)} $ by  Eq. (\ref{train_error}) for  up to $T_e$ iterations,  and count the number of early stopped base neural networks $n_e$ 
    \STATE  Update $X_{\textmd{C}}$ by greedy ensemble search with Eq. (\ref{update_candidate1}) for up to $T_G = \delta_G\lfloor \frac{2n_e}{L}+1\rfloor$ iterations
    \STATE Select state $\hat{\mathbf{x}}_{\textmd{C}} $ by Eq. (\ref{candidate1})
    \STATE $ e_{\textmd{R,C}}= e_{\textmd{R}}(\hat{\mathbf{x}}_{\textmd{C}}), \quad e_{\textmd{F}}= e_{\textmd{F}}(\hat{\mathbf{x}}_{\textmd{C}})$
    \STATE $\hat{\mathbf{x}}^*= \hat{\mathbf{x}}_{\textmd{C}}$
    \STATE \textbf{do line 4-6}
    \STATE $\hat{\mathbf{x}}_{\textmd{M}} =\mathcal{U}[\mathbf{x}_{min}, \mathbf{x}_{max}], \quad e_{\textmd{R,M}}= e_{\textmd{R}}(\hat{\mathbf{x}}_{\textmd{M}} )$
    \STATE $D^{(t)} = D^{(t)}  \cup \left(\hat{\mathbf{x}}_{\textmd{C}}, e_{\textmd{R,M}}\right) \cup\left(\hat{\mathbf{x}}_{\textmd{M}}, e_{\textmd{R,M}}\right) $
    
\ENDFOR
  \end{algorithmic}
  \end{algorithm}

\section{Experiments And Analysis}
We hope ML models can work well in the \textit{close world} described by training dataset, however,  the test samples in the deployment are from the \textit{open world} where diverse samples exist, and differ from the distribution of training set. In this section, we will test the proposed SPEC under four different deployment conditions: (1) Inside Distribution (ID) Test: Test the proposed SPEC with the data from the same distribution of the training set, which is the conventional scenario that people test ML models on; (2) Outside of Distribution (OoD) Test: test the proposed SPEC with the data with the sample outside the distribution of the training set; (3) Samples inconsistented with physical model: test the samples which cannot fully simulated by the physical model $F$, which is often happened in practice because of noise and measurement error. (4)Reconfigurability test: The last test is to exhibit the reconfigurability of the proposed SPEC, that is, the proposed SPEC can be reconfigured to adapt to thoroughly new applications, such as different wavebands, different types of spectral signals, etc.

\subsection{Inside Distribution (ID) Test}
\begin{figure}[hbtp!]
  \centering
  \includegraphics[width=1.\columnwidth]{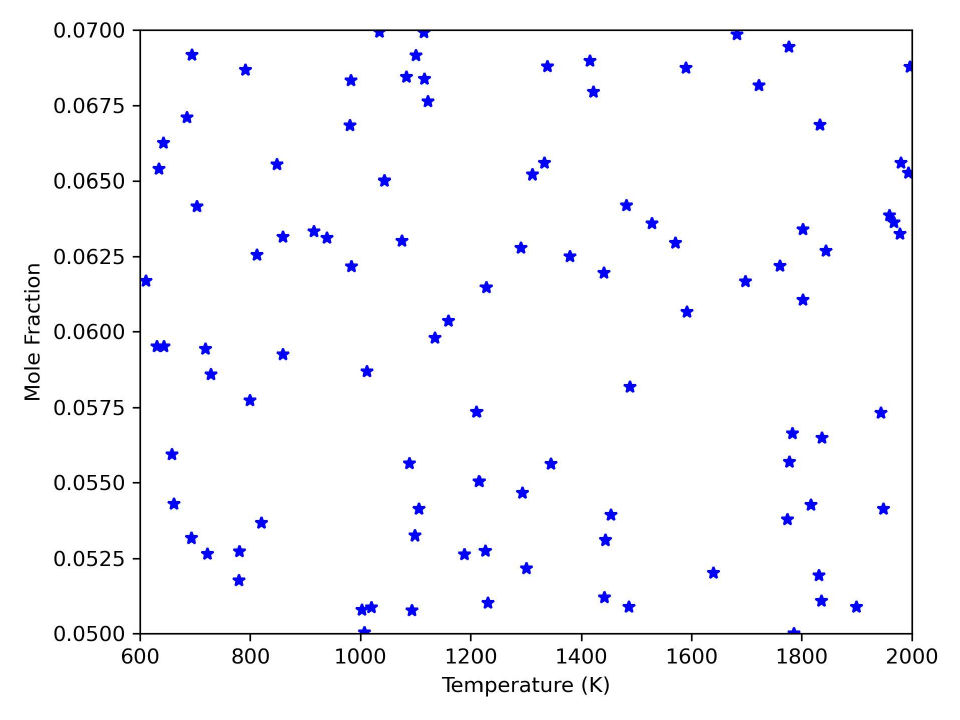}
  \caption{The distribution of the ID test set.}
  \label{fig:id_distribution}
\end{figure}

In very often cases, we prefer the training and test sets satisfy the hypothesis of independent and identical distribution (I.I.D), and we often assess the model under this condition. Such test data does exist in the deployment, so we first test SPEC with 100 samples randomly picked from the test set we acquired along with the training set generation.
The distribution of this tiny test set is shown in Fig. \ref{fig:id_distribution}. Under such case, the feasible domain $\mathcal{X}_\textmd{F}$ is as the same as the data generation range, i.e., temperature range is 600 to 2000 K, mole fraction range is 0.05 to 0.07.

\begin{figure}[hbtp!]
  \centering
  \includegraphics[width=1.\columnwidth]{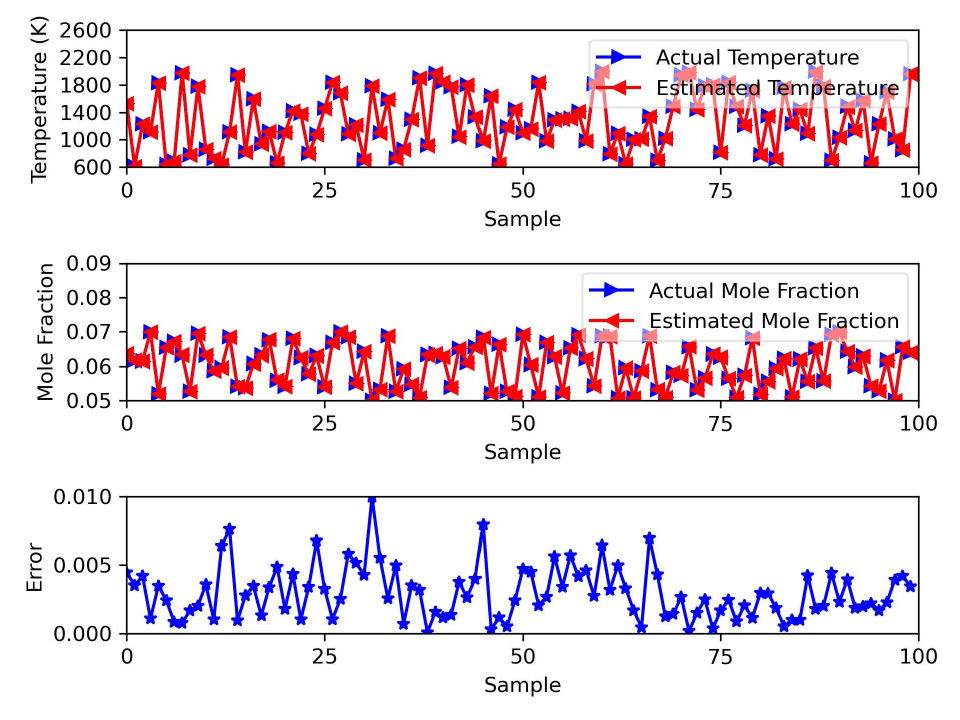}
  \caption{The performance of existed ML estimator (estimation mode) on the ID test set. The upper panel shows the comparison of the temperature estimation of ML estimator $G$ and the ground truth. The middle panel shows the comparison of the concentration estimation of ML estimator $G$  and the ground truth. The lower panel shows the overall error calculated via PAD.}
  \label{fig:ID_result}
\end{figure}

We set the error threshold $\epsilon=0.05$. The test results are shown in Fig. \ref{fig:ID_result}. Because the test set is thoroughly consistent with the training set. Therefore, Estimation Mode, i.e.,  a well-trained ML estimator $G$, can work well on the test set. Visually, the estimation of temperature and concentration are basically overlapped to their ground truth (upper and middle panel in Fig.\ref{fig:ID_result}). However, it is worth noting that the ground truth is not available in the deployment, after all, ML models are designed to estimate the ground truth when ground truth is not available. That is exact the reason why conventional ML methodology lacks the tool to assess their estimation reliability. But PAD offers the route to assess the reliability of estimations without the ground truth. From the calculation of PAD, the highest error is merely 0.01, much smaller than 0.05 (lower panel in Fig.\ref{fig:ID_result}). Therefore, the proposed SPEC does not need to activate the correction mode under this test scenario.

\subsection{Outside of Distribution (OoD) Test}
\begin{figure}[hbtp!]
  \centering
  \includegraphics[width=1.\columnwidth]{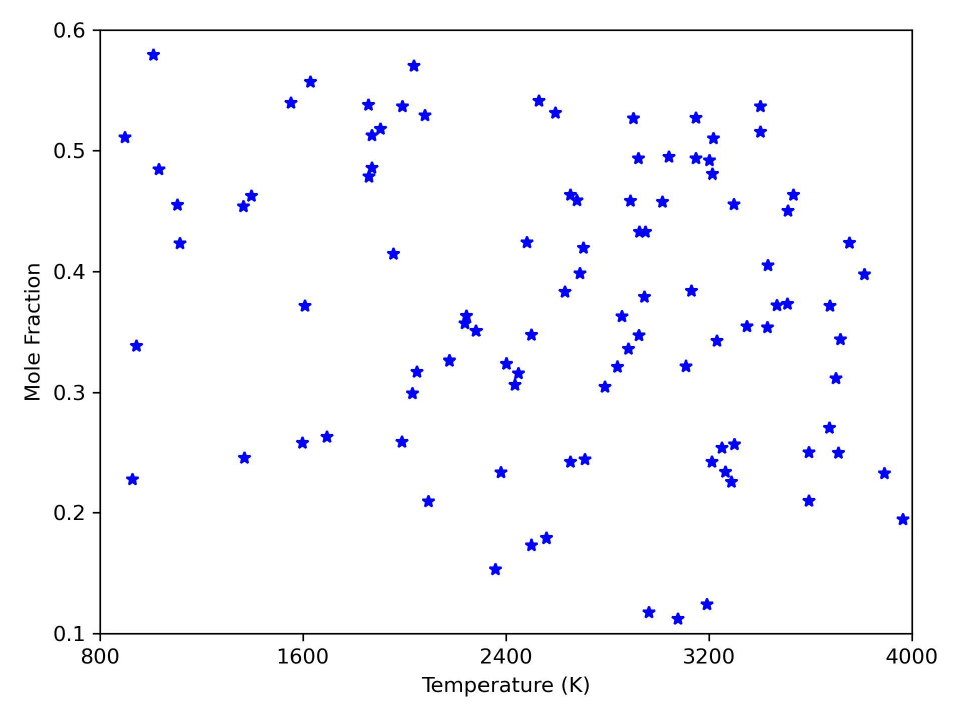}
  \caption{The distribution of the OoD test set.}
  \label{fig:ood_distribution}
\end{figure}

We also test the proposed SPEC with 100 samples outside the distribution of training set. These samples are randomly sampled from the range of temperature between 800 and 4000K, and mole fraction between 0.1 and 0.6. The distribution of the test set is shown in Fig. \ref{fig:ood_distribution}. We accordingly randomly generate the feasible domain for each state.

\begin{figure}[hbtp!]
  \centering
  \includegraphics[width=1.\columnwidth]{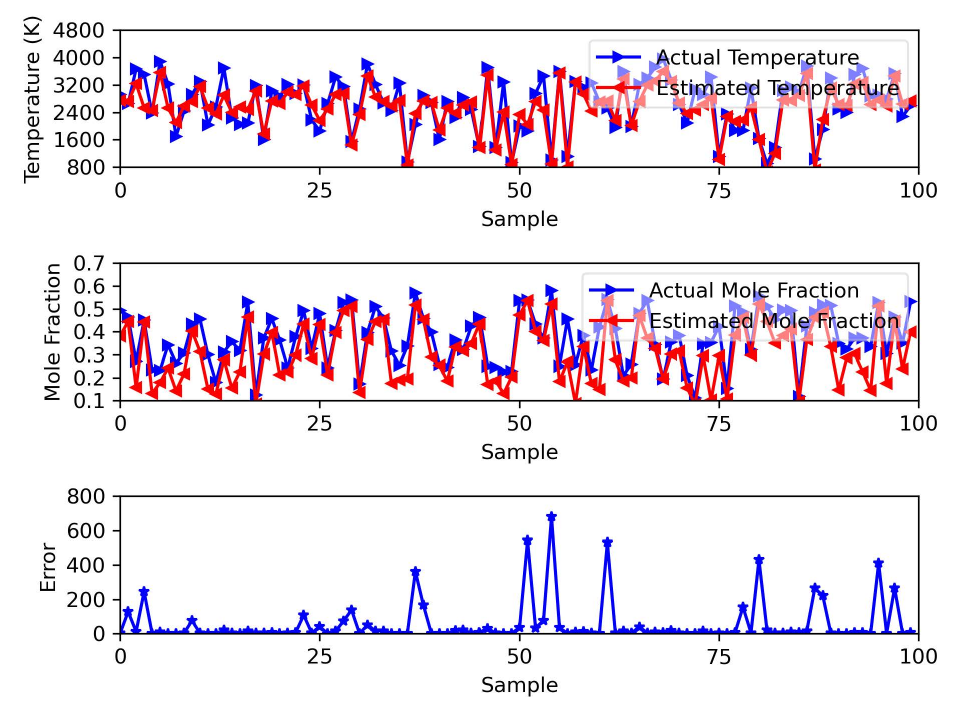}
  \caption{The performance of existing ML estimator (estimation mode) on the OoD test set. The upper panel shows the comparison of the temperature estimation of ML estimator $G$ and the ground truth. The middle panel shows the comparison of the concentration estimation of ML estimator $G$  and the ground truth. The lower panel shows the overall error calculated via PAD.}
  \label{fig:ood_estimator}
\end{figure}

We respectively set the acceptable error threshold $\epsilon=0.05, 0.075, and 0.1$, and start the experiment. Estimation Mode is first activated to provide efficient estimation. The comparison between estimation and ground truth of states is shown in Fig. \ref{fig:ood_estimator}. In general, estimation mode can capture the trend of the ground truth. However, according to the assessment of PAD, the minimal error is 0.373, which is much larger than the most relaxing error threshold $\epsilon= 0.1$, meanwhile, the maximal error is even up to 681.28. This phenomenon indicates the value of PAD and correction mode in the deployment of ML model.

Accordingly, the correction mode of SPEC is activated to correct the estimation. In addition to our proposed correction mode, we also compare to other neural-network-based optimization algorithms, including SVPEN \cite{kangSelfValidatedPhysicsEmbeddingNetwork2022},Tandem Network \cite{guan2023machine}, Dummy Input Layer \cite{peurifoyNanophotonicParticleSimulation2018a}, and PSO-Network Hybrid Model \cite{vaezinejadHybridArtificialNeural2019}. Two metrics are used in this test, they are respectively the failure times and Average iteration. The former is  defined as the failure times of the model to meet the error threshold in these 100 test cases, and the latter is defined as the average number of iterations needed to find the acceptable state. 

The results are shown in Table \ref{tab:ood_result}, which tells that SPEC outperforms other algorithms in both metrics of failure times and the average iteration times. The reason behind is that SPEC has robust and efficient search capability compared to other algorithms. Although starting from one state, fine-tuning or training a network model is workable (SVPEN, Tandem Network, Dummy Input Layer), they are easily to be trapped into local minima, while directly searching surrogate model faces the risk of overfitting.

\begin{table*}[]
  \centering
  \caption{Algorithm performance on OoD test set under different thresholds (optimal results are marked in \textbf{bold}, and the second best results are \ul{underlined}.)}
  \label{tab:ood_result}
  \resizebox{0.9\textwidth}{!}{%
  \begin{tabular}{|c|cc|cc|cc|}
  \hline
  Threshold      & \multicolumn{2}{c|}{0.05}         & \multicolumn{2}{c|}{0.075}        & \multicolumn{2}{c|}{0.1}          \\ \hline
  Method         & Failure times & Iteration         & Failure times & Iteration         & Failure times & Iteration         \\ \hline
  SVPEN          & 29            & 116.83 \textpm 43.35      & 15            & 100.10 \textpm 43.30      & 7             & 98.46 \textpm 38.71       \\
  Tandem Network & {\ul 18}      & 109.36 \textpm 40.57      & {\ul 6}       & 94.66 \textpm 37.79       & {\ul 1}       & 83.80 \textpm 33.59       \\
  Dummy input Layer          & 19           & {\ul73.83 \textpm 56.98}      & 11           & {\ul 56.78 \textpm 51.30}       & 6           & 47.77 \textpm 45.86     \\
  PSO-Network Hybrid Model         & 37            &  83.96 \textpm 91.13 & 25            &  57.22  \textpm 83.43 & 13            & {\ul 31.18 \textpm 65.69} \\
  Proposed SPEC & \textbf{7} & \textbf{30.33 \textpm 54.74} & \textbf{2} & \textbf{15.57 \textpm 39.03} & \textbf{0} & \textbf{7.63 \textpm 20.28} \\ \hline
  \end{tabular}%
  }
\end{table*}

\begin{figure}[hbtp!]
  \centering
  \includegraphics[width=1.\columnwidth]{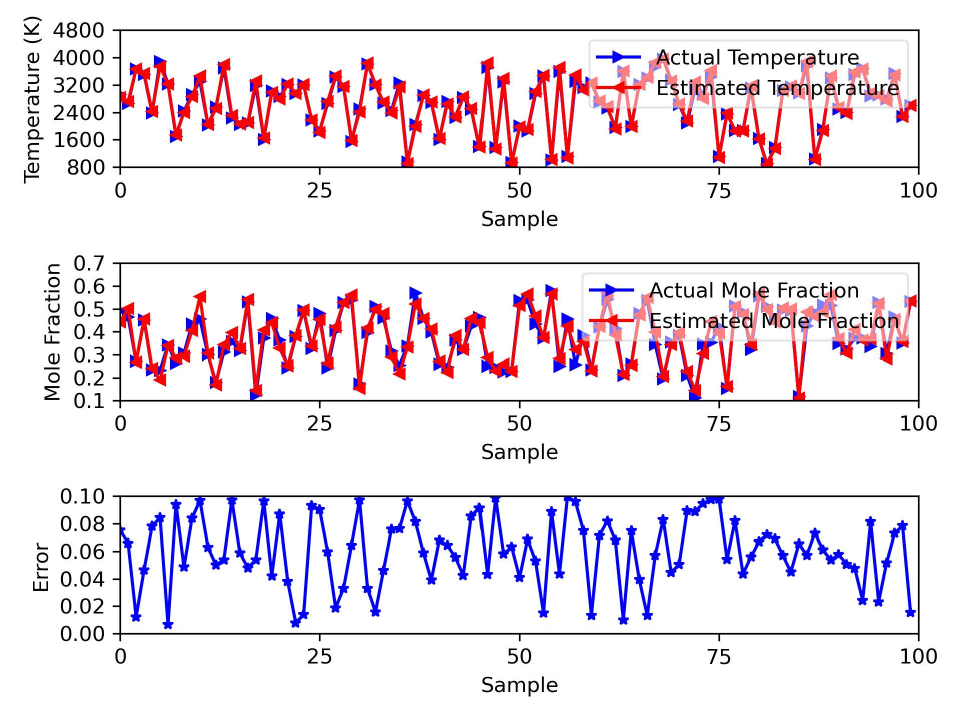}
  \caption{The correction performance on OoD test set under the threshold of 0.1. The upper panel shows the comparison of the corrected temperature estimation of SPEC and the ground truth. The middle panel shows the comparison of the corrected concentration estimation of SPEC and the ground truth. The lower panel shows the overall error of corrected state calculated via PAD.}
  \label{fig:ooD_result}
\end{figure}

Figure\ref{fig:ooD_result} shows the corrected states under the error threshold of 0.1. Compared to Fig.\ref{fig:ood_estimator}, the corrected estimation is much closer to the ground truth. The overall error is also significantly reduced. The results tell that SPEC can effectively correct the unreliable estimation of ML estimator.

A quantitative comparison between estimation and correction modes is done. In this comparison, we calculate the Root Mean Square Error (RMSE), Mean Absolute Error (MAE), Mean Relative Error (MRE), and Pearson Correlation Coefficient (R) of the corrected estimation and ground truth. The results are shown in Table \ref{tab:ood correction benefit}. The results tell that after activating correction mode with a threshold of 0.05, the RMSE, MAE, RE and R on temperature are respectively improved by 80.6\%, 82.2\%, 82.7\% and 0.079. On concentration, the RMSE, MAE, RE and R are respectively improved by 69.4\%, 71.4\%, 70.8\% and 0.068. When the threshold is even more critical to 0.05, the RMSE, MAE, RE and R on temperature are respectively improved by 87.6\%, 88.2\%, 88.3\% and 0.087. On concentration, the RMSE, MAE, RE and R are respectively improved by 78.0\%, 79.0\%, 78.8\% and 0.076.  Especially, the relative error are reduced to single-digit orders of percentage, which is sufficient in most engineering applications. 

\begin{table*}[hbtp!]
  \centering
  \caption{The benefits of correction mode on OoD test set under different thresholds.}
  \label{tab:ood correction benefit}
  \resizebox{0.75\textwidth}{!}{%
  \begin{tabular}{ccccccccc}
  \hline
  State Elements  & \multicolumn{4}{c}{Temperature}   & \multicolumn{4}{c}{Concentration} \\ \hline
  Metric         & RMSE    & MAE     & MRE    & R      & RMSE    & MAE     & MRE    & R      \\ \hline
  Estimation Mode & 337.332 & 264.954 & 0.104 & 0.918 & 0.093  & 0.073  & 0.211  & 0.894  \\
  $\epsilon=0.1$             & 65.546  & 47.206  & 0.018 & 0.997 & 0.028  & 0.021  & 0.063  & 0.973  \\
  $\epsilon=0.075$          & 51.441  & 38.915  & 0.015 & 0.998 & 0.022  & 0.018  & 0.052  & 0.982  \\
  $\epsilon=0.05$            & 41.814  & 31.022  & 0.012 & 0.999 & 0.017  & 0.014  & 0.041  & 0.990  \\ \hline
  \end{tabular}%
  }
  \end{table*}
\subsection{Samples inconsistented with physical model}
\label{sec:inconsistent}
Real-world phenomena are not often exactly same as the simulation via physical model due to the simulation fidelity limitation, measure error, etc. Therefore, for a given state $\mathbf{x}$, the spectrum measured in actual deployment may be different from the spectrum simulated by physical model. In this case, the inconsistency between test data and physical model is introduced, i.e.,
\begin{equation}
  \mathbf{y} = \mathbf{G}(\mathbf{x}) + \Delta \mathbf{y} ,
\end{equation}
where, $\Delta \mathbf{y}$ is the inconsistency between test data and physical model.

In this experiment, we use 10\% Gaussian noise to be the inconsistency, and add it to the ID test data to simulate the actual measurement $\mathbf{y}$, which is defined as:
\begin{equation}
  \mathbf{y}[i] = \mathbf{G}(\mathbf{x})[i] (1+0.1\mathcal{N}(0,1)), \quad i \in [1, |\mathbf{y}|].
\end{equation}

\begin{figure}[hbtp!]
  \centering
  \includegraphics[width=1.\columnwidth]{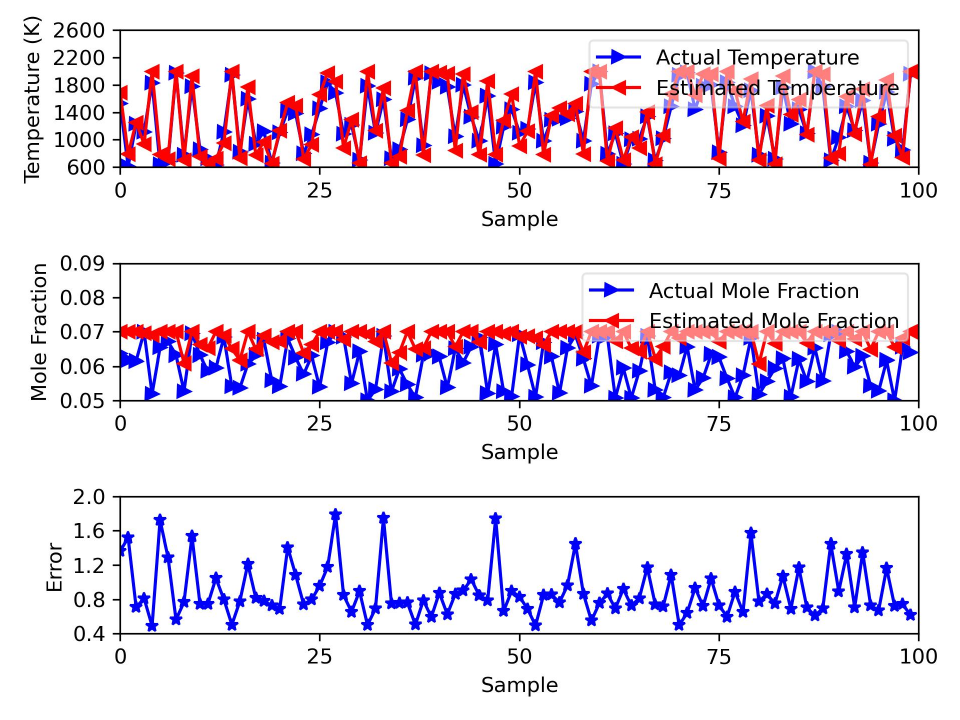}
  \caption{The performance of existing ML model on the noise-added ID test set. The upper panel shows the comparison of the temperature estimation of ML estimator $G$ and the ground truth. The middle panel shows the comparison of the concentration estimation of ML estimator $G$  and the ground truth. The lower panel shows the overall error calculated via PAD.}
  \label{fig:inconsistency_result_estimation}
\end{figure}

We set the threshold to be 0.05. Because of the existence of inconsistency, the estimation mode cannot provide accurate estimation. As shown in Fig.\ref{fig:inconsistency_result_estimation},  the estimation of temperature is fine but the estimation of concentration is visually inaccurate. The minimal overall error is more than 0.4, and maximal error is  even about 1.8. Such an estimation peformance is intolerable, and thus the correction mode is activated.

\begin{figure}[hbtp!]
  \centering
  \includegraphics[width=1.\columnwidth]{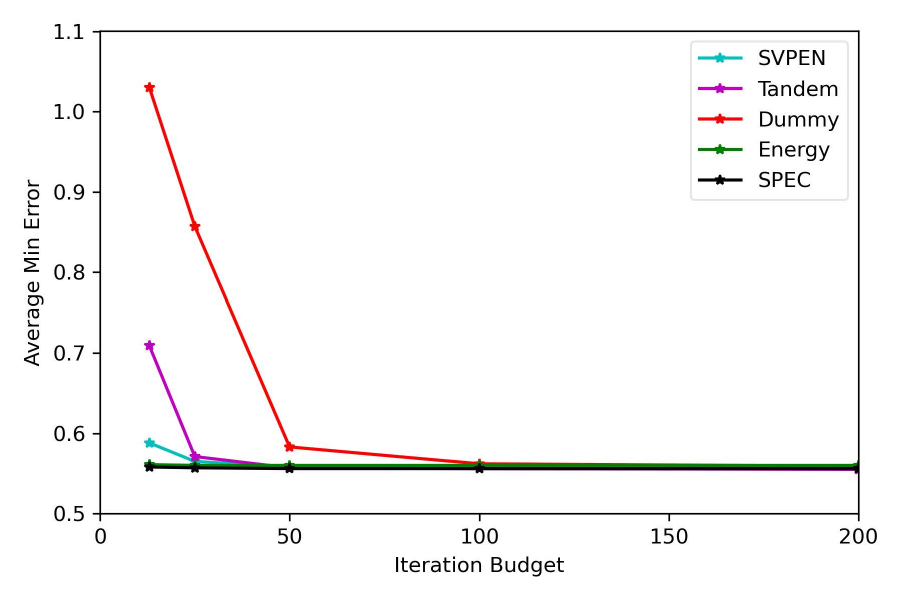}
  \caption{The performance of ML on the noise-added ID test set.}
  \label{fig:noise_result}
\end{figure}

Meanwhile, because of the existence of the inconsistency, the ideal error measured by PAD is not zero any more, but it is unknown to us. Therefore, the predefined error threshold could be inappropriate. Therefore, under such condition, we use the average lowest error $e_{min}$ in limited iteration budget $T$ to judge algorithms. In this experiment, the iteration budget $T$ is defined as 25, 50,100, and 200. The results shown in Fig.\ref{fig:noise_result} tells that all algorithms converge to the similar level of error, but the proposed SPEC converges faster than other algorithms, which is benefited from the proposed greedy ensemble search.

\begin{figure}[hbtp!]
  \centering
  \includegraphics[width=1.\columnwidth]{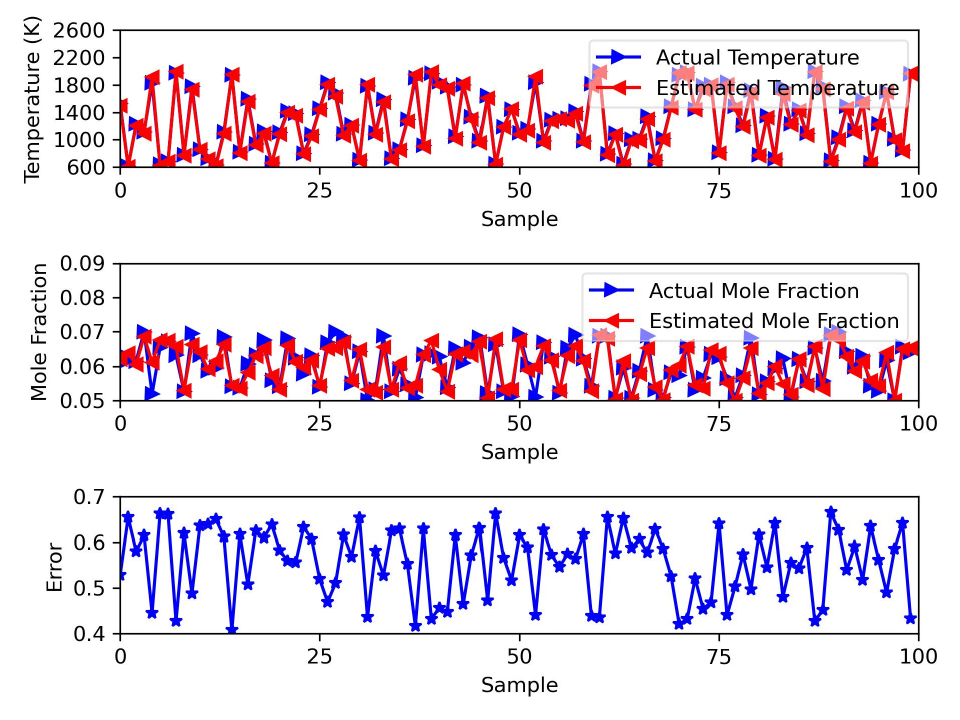}
  \caption{The performance of correction mode on the noise-added ID test set. The upper panel shows the comparison of the temperature estimation of the correction mode and the ground truth. The middle panel shows the comparison of the concentration estimation of  the correction mode  and the ground truth. The lower panel shows the overall error calculated via PAD.}
  \label{fig:inconsistency_result_correction}
\end{figure}

We plot the corrected estimation of SPEC after 200 iterations into Fig.\ref{fig:inconsistency_result_correction}. Compared to Fig.\ref{fig:inconsistency_result_estimation}, we can visually observe that the estimation of concentration is significantly improved. The maximal overall error is also reduced from 1.8 to no more than 0.7.

More quantitative assessment is provided in Table \ref{tab:noise correction benefit}. Similar to the result shown in Table \ref{tab:ood correction benefit}, the results tell that the correction mode can significantly improve the estimation accuracy. After activating correction mode with a threshold of 0.05, the RMSE, MAE, RE and R on temperature are respectively improved by 80.0\%, 82.3\%, 83.7\% and 0.022, and the RMSE, MAE, RE and R on concentration are respectively improved by 80\%,75\%, 81.1\% and 0.604. Of course, one can design more capable estimator $G$ to be robust for noise, which is welcomed in practice and not conflict with the proposed SPEC, the emphasis of the results herein is to demonstrate the capability of correction mode.

\begin{table*}[]
  \centering
  \caption{The benefits of correction mode on noise-added ID test set.}
  \label{tab:noise correction benefit}
  \resizebox{0.75\textwidth}{!}{%
  \begin{tabular}{ccccccccc}
  \hline
  State Elements  & \multicolumn{4}{c}{Temperature}   & \multicolumn{4}{c}{Concentration} \\ \hline
  Metric          & RMSE    & MAE     & MRE   & R     & RMSE   & MAE    & MRE    & R      \\
  Estimation Mode & 133.485 & 117.647 & 0.098 & 0.977 & 0.010  & 0.008  & 0.148  & 0.327  \\
  Correction Mode $T=200$ & 26.777  & 20.831  & 0.016 & 0.999 & 0.002  & 0.002  & 0.028  & 0.931  \\ \hline
  \end{tabular}%
  }
  \end{table*}

\subsection{Reconfigurability of SPEC}
\begin{figure*}[hbtp!]
  \centering
  \includegraphics[width=0.8\textwidth]{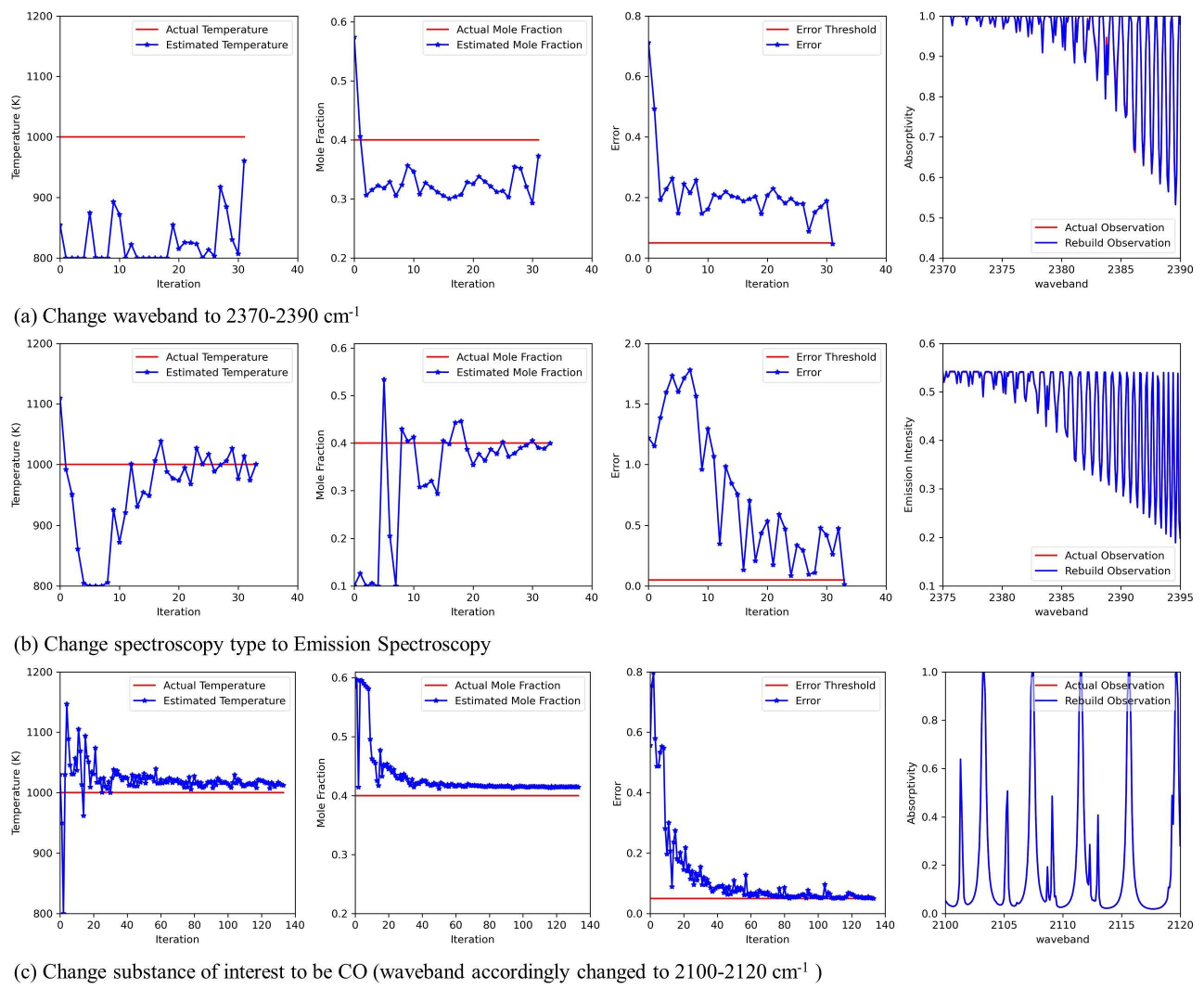}
  \caption{The reconfigurability of SPEC, where we use (a) the absorption spectrum from another waveband, (b) the emission spectrum instead of absorption spectrum; (c)  The absorption spectrum of CO from another waveband, and even  as the test data, and change to their corresponding physical model in SPEC. In each subfigure, four panels from left to right are respectively the iteration process of temperature, concentration, overall error, and the comparison of the test spectrum and the rebuild spectrum.}
  \label{fig:reconfigurability_result}
\end{figure*}

Compared to the conventional ML-solutions in spectroscopy quantification, the added correction mode in SPEC makes it reconfigurable, i.e., through changing physical forward model embedded in PAD, SPEC can tackle different types of spectra.
Some examples are shown in Fig.\ref{fig:reconfigurability_result}, where we use the absorption spectrum from another waveband, absorption spectrum from another substance (CO), and even the emission spectrum as the test samples. Without doubt, these cases are outside the capability of estimation mode as well as conventional ML models, because their knowledge is fixed in training set, while these samples are thoroughly different from the training set (Comparing Fig.\ref{fig: waveband sensitivity} and Fig.\ref{fig:reconfigurability_result}). Fortunately, benefited from the introduction of correction mode, through changing the physical model to the corresponding configuration, SPEC can still achieve a reasonable estimation result.

This property is very useful in practice, because when we deploy the model into actual applications, the waveband sometimes can be not exactly same as the one for training because of the measurement error. And sometimes, we realize the knowledge in existing ML estimator cannot handle needed application scenarios. In conventional ML solutions, we often need to retrain the ML estimator, but because of the introduction of correction mode, we do not fall into such trouble, but just change the physical model embedded in PAD. In addition, this kind of setting also provides the users the flexibility to update or change the physical model at willing.

\section{Ablation Study}
We mainly did three ablation studies herein: (1) The effect of ways to estimate error; (2) The effect of sampling methods; (3) The effect of diversity error.

\textbf{The effect of ways to estimate error}. In this study, we respectively use base networks to estimate all three error elements, overall error, and merely reconstruction error. The results shown in Table \ref{tab:error estimation} tells that merely estimating reconstruction error can achieve the best performance. The reason is that estimating all error elements will add more estimation uncertainty and thus confuse the optimization direction, while estimating the overall error will lose the information of error structure. Notably, the difference between estimating overall error and estimating reconstruction error is limited, this is because our states are sampled from the feasible domain at the beginning, thus, very few states will activate the feasible error, therefore, the overall error is mainly affected by the reconstruction error.

\begin{table}[]
  \centering
  \caption{ The effect of ways to estimate error}
  \label{tab:error estimation}
  \resizebox{\columnwidth}{!}{%
  \begin{tabular}{ccc}
  \hline
  Ways of estimating error                    & Failure times & Iteration      \\ \hline
  Estimate all error elements & 19                 & 56.72  \textpm 77.42          \\
  Estimate overall error        & 8                  & 30.88 \textpm  56.92          \\
  Estimate reconstruction error & \textbf{7}         & \textbf{30.33 \textpm 54.74} \\ \hline
  \end{tabular}%
  }
  \end{table}

\textbf{The effect of sampling methods}. In this study, we compare the used simple Monte-Carlo sampling with active sampling via disagreement \cite{shyamModelBasedActiveExploration2019}. The latter is an active sampling method which uses a ML model to find the states lead to the maximal disagreement between the estimation of base networks. The logic behind this method is that the states with maximal disagreement are the most uncertain states which need to enhance.
The results shown in Table \ref{tab:exploration} tells that the active sampling method is not suitable for SPEC, because the most uncertain state does not mean the most important state. In fact, active sampling will frequently shift the focus of surrogate model and accordingly make the optimization process unstable. This is also indicated by the high standard deviation of active sampling method in Table \ref{tab:exploration}.
\begin{table}[]
  \centering
  \caption{The effect of different sampling methods}
  \label{tab:exploration}
  \resizebox{\columnwidth}{!}{%
  \begin{tabular}{ccc}
  \hline
  Sampling method                  & Failure times & Iteration      \\ \hline
  Active sampling & 25                 & 59.48  \textpm 83.46          \\
  Monte-Carlo sampling & \textbf{7}         & \textbf{30.33 \textpm 54.74} \\ \hline
  \end{tabular}%
  }
  \end{table}

\textbf{The effect of diversity error}. In this study, we compare cases of using diversity error and not using diversity error in greedy ensemble search (Eq.\ref{update_candidate1}). The results shown in Table \ref{tab:diversity error} tells that using diversity error reduces the failure times and iteration standard deviation. The reason is that the diversity error can help the ensemble search to avoid the local optimum, and thus improve the stability of the optimization process.

  \begin{table}[]
    \centering
    \caption{The effect of diversity error}
    \label{tab:diversity error}
    \resizebox{\columnwidth}{!}{%
    \begin{tabular}{ccc}
    \hline
    diversity error                  & Failure times & Iteration      \\ \hline
    w/o diversity error & 9                & 30.50  \textpm 59.96        \\
    with diversity error & \textbf{7}         & \textbf{30.33 \textpm 54.74} \\ \hline
    \end{tabular}%
    }
    \end{table}

\section{Discussion}
Although the proposed SPEC has achieved promising results, it has some aspects can be improved. The correction mode is a network-based optimization algorithm in nature. Therefore, before it becomes workable, it needs to collect same data to activate the training of surrogate model, although this pre-collected data amount is very few in our algorithm. To complement this warm-up period, one can treat the Monte-carlo sampling as an independent optimization route, and take the explored state it collects as the searched state to be compared with the error threshold. Such a complementation does not only cover the warm-up period of correction mode, but also provide an unbiased search route to the users, whereas current network-based optimization is biased by the distribution of collected data.

In fact, one can enhance the correction mode by blending different optimization algorithms , for example, Particle Swarm Optimization (PSO), Genetic Algorithm (GA), Bayesian Optimization, etc. However, it is notable that any added optimization algorithm will increase the computational cost of correction mode, which is not desirable in practice. In addition, these algorithms
also have their own limitations. GA and PSO need to query PAD very often to evaluate the quality of state, while many physical forward modelling platforms are time-consuming \cite{kochanovHITRANApplicationProgramming2016}. Accordingly, thousands of sampling and querying may be unaffordable.
The time needed for Bayesian optimzation is increasing exponentially with the iteration times. For the case that optimization has to go through a fixed huge iteration budget, such as case in Sec. \ref{sec:inconsistent}, it will become very time-consuming. Compared to them, Monte-carlo optimization is a cheaper choice, as it has been combined inside the correction mode for data exploration. How to select and combine different algorithms inside correction mode is one important direction as our future work.

In our work, we merely discussed the case of single substance, however, it is very common that the test data is a mixture of multiple substances. In this case, the state dimension may be pretty high, so one may need to consider more efficient exploration instead of simple Monte-Carlo sampling, this is also an important direction as our future work. By the way, under such condition, Bayesian optimization we mentioned above may be also not suitable, because its performance degrades quickly when the state dimension is high \cite{tran-the2020trading}.

At last, we have to emphasize that current SPEC does not have any conflict with the endeavors of making ML estimators powerful. In fact, such endeavors are welcomed. Because a better ML estimator will decrease the work load of proposed correction mode, and thus improve the overall performance of SPEC. 

\section{Conclusion}
In this work, we proposed a novel framework, SPEC,  to address the reliability issue of spectroscopy quantification. The proposed SPEC is a dual-mode framework containing estimation and correction modes. The estimation mode is a conventional ML estimator, which is trained via the precollected data set, and has the function to estimate temperature and concentration of substance (state ) according to the given spectrum. For an estimation provided by estimation mode, a Physics-driven Anomaly Detection (PAD) module is constructed to assess its quality indirectly. If the estimation from estimation mode cannot pass the assessment. The correction mode will be activated to correct the estimation. The correction  mode is a network-based optimization algorithm, which utilizes the error information feedback from PAD to guide the optimization. From above experiments and analysis, we can conclude following points:
\begin{itemize}
  \item SPEC is a harmonious combination of ML and physical model. The physics-Driven Anomaly Detection Module does not only provide reliable detection of ML estimation, but also provide accurate optimization guidance to correct the estimation.
  \item SPEC combines the ML estimator and also the Network-based optimization algorithm into a unified framework, which provides the ML-based estimation the self-correction capability. The proposed algorithm for correction mode outperforms current network-based optimization algorithms in terms of the success rate and efficiency.
  \item The proposed SPEC demonstrates its effectiveness in test data that inside or outside the training data distribution, as well as the noisy test data. In addition, it has the property of reconfigurability, which can be easily adapted to different types of spectra by changing the configuration of PAD.
\end{itemize}

\section*{Acknowledgments}
Kang R. would like to acknowledge Profs. Dimitrios Kyritsis and Tiejun Zhang at Khalifa University for their instruction on spectroscopy quantification.

\bibliographystyle{IEEEtran}
\bibliography{reference}

\newpage

 




\vfill

\end{document}